\documentclass[conference]{IEEEtran}

\usepackage{algorithmic}
\usepackage{amsmath,amssymb,amsfonts}
\usepackage{balance}
\usepackage{bm}
\usepackage{cite}
\usepackage{colortbl}
\usepackage[T1]{fontenc}
\usepackage{graphicx}
\usepackage{makecell}
\usepackage{mathtools}
\usepackage{multirow}
\usepackage{siunitx}
\usepackage{soul}               % for highlight
\usepackage{textcomp}
\usepackage{threeparttable}
\usepackage[color=green!40]{todonotes}
\soulregister\cite7
\soulregister\footnote7
\soulregister\eqref7
\soulregister\ref7
\usepackage{xcolor}
\usepackage{hyperref}
\graphicspath{{./}{figures/}{../figures/}}

\begin{document}

\title{%
  Hierarchical Multi-Building And Multi-Floor Indoor Localization Based On
  Recurrent Neural Networks%
}

\author{%
  % \IEEEauthorblockN{Authors names are omitted for double blind review}%
  \IEEEauthorblockN{Abdalla Elmokhtar Ahmed Elesawi and Kyeong Soo Kim}%
  % \IEEEauthorblockA{Authors affiliations are omitted for double blind review}%
  \IEEEauthorblockA{%
    % Department of Communications and Networking\\%
    School of Advanced Technology, Xi'an Jiaotong-Liverpool University, Suzhou,
    215123, P. R. China.\\%
    Email: A.Elesawi19@student.xjtlu.edu.cn, Kyeongsoo.Kim@xjtlu.edu.cn%
  }%
}%

\maketitle

\begin{abstract}
  There has been an increasing tendency to move from outdoor to indoor lifestyle
  in modern cities. The emergence of big shopping malls, indoor sports
  complexes, factories, and warehouses is accelerating this tendency. In such an
  environment, indoor localization becomes one of the essential services, and
  the indoor localization systems to be deployed should be scalable enough to
  cover the expected expansion of those indoor facilities. One of the most
  economical and practical approaches to indoor localization is Wi-Fi
  fingerprinting, which exploits the widely-deployed Wi-Fi networks using mobile
  devices (e.g., smartphones) without any modification of the existing
  infrastructure. Traditional Wi-Fi fingerprinting schemes rely on complicated
  data pre/post-processing and time-consuming manual parameter tuning. In this
  paper, we propose hierarchical multi-building and multi-floor indoor
  localization based on a recurrent neural network (RNN) using Wi-Fi
  fingerprinting, eliminating the need of complicated data pre/post-processing
  and with less parameter tuning. The RNN in the proposed scheme estimates
  locations in a sequential manner from a general to a specific one (e.g.,
  building${\rightarrow}$floor${\rightarrow}$location) in order to exploit the
  hierarchical nature of the localization in multi-building and multi-floor
  environments. The experimental results with the UJIIndoorLoc dataset
  demonstrate that the proposed scheme estimates building and floor with 100\%
  and 95.24\% accuracy, respectively, and provides three-dimensional positioning
  error of 8.62\,m, which outperforms existing deep neural network-based
  schemes.
\end{abstract}


\section{Introduction}
\label{sec:introduction}
In modern smart cities, there is a huge demand for location-based services (LBS)
like advertising, tracking, and navigation. Because people spend most of their
time in indoor environments like shopping malls, hospitals, and
airports~\cite{osti_6958939}, we need to provide LBS indoors. Most existing
localization systems, however, cannot be used indoors; the lack of line of sight
and the effect of multipath propagation on signals make it hard to utilize the
localization systems such as the global positioning system (GPS) for indoor
localization~\cite{bahl00:_radar}.

Different technologies have been utilized to implement indoor LBS (ILBS),
including wireless networks, active/passive tags, and vision/camera
technologies~\cite{basri:survey}. Among many wireless technologies for indoor
localization
% such as Wi-Fi, Bluetooth, ZigBee, RFID, Ultra-Wideband (UWB) and ultrasound,
Wi-Fi is the most feasible and popular technology~\cite{zaf:survey};
% Wi-Fi networks are widely available, ready to use, and supported by
% most of mobile devices.
%
% Angle of arrival (AOA), time of arrival (TOA), and time difference of arrival
% (TDOA) are some of the major localization techniques, which, however, require
% the complete and accurate information on network topologies such as the
% locations of receivers and/or transmitters and the orientations of
% antenna~\cite{chris:survey}.
Especially, Wi-Fi fingerprinting is widely used for indoor localization due to
its wide availability and the lack of strict requirements on the information on
network topologies. In Wi-Fi fingerprinting, received signal strength (RSS)
measurements are not directly used for distance estimation with a path loss
model, which is a basis for multilateration, due to multipath fading
effects~\cite{1047316}; instead, Wi-Fi fingerprinting uses RSS as one of
location-dependent characteristics (i.e., location fingerprint) in inferring the
location. We first build a database of the RSS indicators (RSSIs) from all
access points (APs) measured at known locations called reference points (RPs)
together with their location information like two dimensional (2D) or three
dimensional (3D) positions. This information should be collected many times for
each RP using different devices, different users, and different orientations to
mitigate the effect of fluctuations in RSS measurements.

Traditional approaches for Wi-Fi fingerprinting---e.g., K-nearest neighbor
(KNN)~\cite{TORRESSOSPEDRA20159263}, weighted KNN (wKNN), and support vector
machine (SVM)~\cite{Amira:svm}---require a lot of efforts for filtering and
parameter tuning, which are quite time-consuming.
% Also, due to the fact that these algorithms are shallow, they can't extract
% very useful complex features.
In recent years, deep neural networks (DNN) have been widely adopted to deal
with large-scale, noisy Wi-Fi fingerprinting
datasets~\cite{nowicki17:_low_wifi,Kim:18-1}, and different machine learning
techniques are combined with DNNs, too~\cite{9262763}. Due to its higher
accuracy and less computational complexity, convolutional neural network (CNN)
are used in~\cite{ibrahim:cnntimeseries,9060340}.

In this paper, we introduce a new approach to hierarchical multi-building and
multi-floor indoor localization based on a recurrent neural network (RNN) with
stacked auto-encoder (SAE) using Wi-Fi fingerprinting, eliminating the need of
complicated data pre/post-processing and with less parameter tuning. The RNN in
the proposed scheme estimates locations in a sequential manner---i.e.,
building${\rightarrow}$floor${\rightarrow}$location---to exploit the
hierarchical nature of the localization in multi-building and multi-floor
environments.

The outline of the rest of the paper is as follows:
Section~\ref{sec:related-work} reviews the related work on indoor
localization. Section~\ref{sec:net-architecture} discusses the proposed network
architecture. In Section~\ref{sec:experimental-results}, we present experimental
results with the best configuration, where we also discuss the results in
comparison with other approaches. Finally, we conclude our work in
Section~\ref{sec:conclusions}.

\section{Related Work}
\label{sec:related-work}
%%%
% In a multi-building and multi-floor environment, 2D coordinates are not enough
% to locate a target; we also need 3D coordinates or the building and the floor
% numbers together with 2D coordinates.
% To achieve accurate localization, different approaches were used.
Traditional approaches like KNN \cite{knn} and wKNN algorithms are time
consuming and need a lot of tuning, which are not suitable for large-scale
indoor environments where a lot of data are to be collected and
processed. Recently, researchers adopt deep learning approaches for indoor
localization.

As for multi-building and multi-floor indoor localization datasets, this work is
based on the publicly-available UJIIndoorLoc
dataset~\cite{torres-sospedra14:_ujiin}. Note that, however, there are several
works using only a subset of
UJIIndoorLoc~\cite{ibrahim:cnntimeseries,Alitaleshi:extermML} or using only
training data for both training and testing their
models~\cite{souad:saeindoor}. Even though these works report good results,
there is no guarantee that the performance of their proposed models would be
good as well with the whole dataset due to the statistical differences in the
training and validation datasets.
% Also, selecting subset of UJIIndoorLoc, most of the time will select good
% records and eliminate outliers, and the model must be evaluated on these
% aspects to see its performance.
In the following, therefore, we focus only on the related works based on the
full UJIIndoorLoc dataset.

In \cite{nowicki17:_low_wifi}, a DNN model was proposed for the classification
of building/floor. An SAE is used to reduce the number of features, which is
followed by a DNN classifier for the \textit{multi-class classification} of
building-floor using flattened labels. Because a flattened label is represented
with one-hot encoding, this DNN model only classifies building and floor to
avoid huge number of output nodes required for locations. This work achieves the
success rate of 92\%.

To tackle the problem of scalability, \textit{multi-label classification} was
proposed in \cite{Kim:18-1}. It reduces the number of output nodes
significantly. The proposed architecture also consists of an SAE followed by a
DNN, but the first $N$ output nodes of the DNN are used for building
classification, where $N$ is the number of buildings, the next $M$ output nodes
for floor classification, where $M$ is the maximum number of floors in all
buildings, and the rest of the output nodes for location estimation. This work
achieves 91.27\% for floor hit rate and \SI{9.29}{\m} for 3D positioning error.

In \cite{9262763}, random forest followed by an SAE was used to filter and
reduce the dimensionality of the dataset. Filtered data is classified using 4
primary classifiers (i.e., CNN, ELM, SVM, and XGBoost), and then the secondary
classifier predicts the class from those 4 values. This work is only for floor
classification, and it achieves 95.13\% for floor hit rate.

Integration of an SAE and CNNs was presented in \cite{9060340}. Three different
networks are used for floor classification, building classification and position
estimation. The SAE followed by dropout layers gives the input to one
dimensional CNN (1D-CNN) followed by fully-connected layers to give 5 output
nodes for floor classification. For building classification, the SAE is directly
connected to fully-connected layers with 3 output nodes. For position
estimation, they used the same floor classification model with some changes:
First, they remove dropout layers between the SAE and the 1D-CNN. Second, they
change the number of output nodes to 2 to represent $x$ and $y$
coordinates. Finally to get continuous values for $x$ and $y$, they use
rectified linear unit (ReLU) instead of softmax as output activation
function. Before training the model, however, lots of data pre-processing is
needed for this work; dividing the dataset to sub-datasets, creating rectangle
areas then dividing them to cell grids, choosing the center of each cell grid,
selecting data based on the previous divisions are some of the preparation steps
for training phase. They achieve 96.03\% for floor hit rate and \SI{11.78}{\m}
for positioning error.

\section{Proposed Network Architecture}
\label{sec:net-architecture}
%%% 
% In this section, we explain the framework for the proposed scheme, starting with
% the structure of a DNN model and its configuration. Then we discuss how to deal
% with the issue of scalability.

% \subsection{Network Architecture}
% %%%
Fig.~\ref{fig:net-architecture} shows the proposed network architecture based on
RNN and SAE, which takes RSSIs as inputs and returns building ID, floor ID, and
location coordinates $(x,y)$ as outputs, where we take into account the
following major points in our design:
%%%
\begin{figure}[!tb]
  \begin{center}
    % \includegraphics[width=.8\linewidth]{rnn1}\\
    % {\scriptsize (a)}\\
    \includegraphics[width=.7\linewidth]{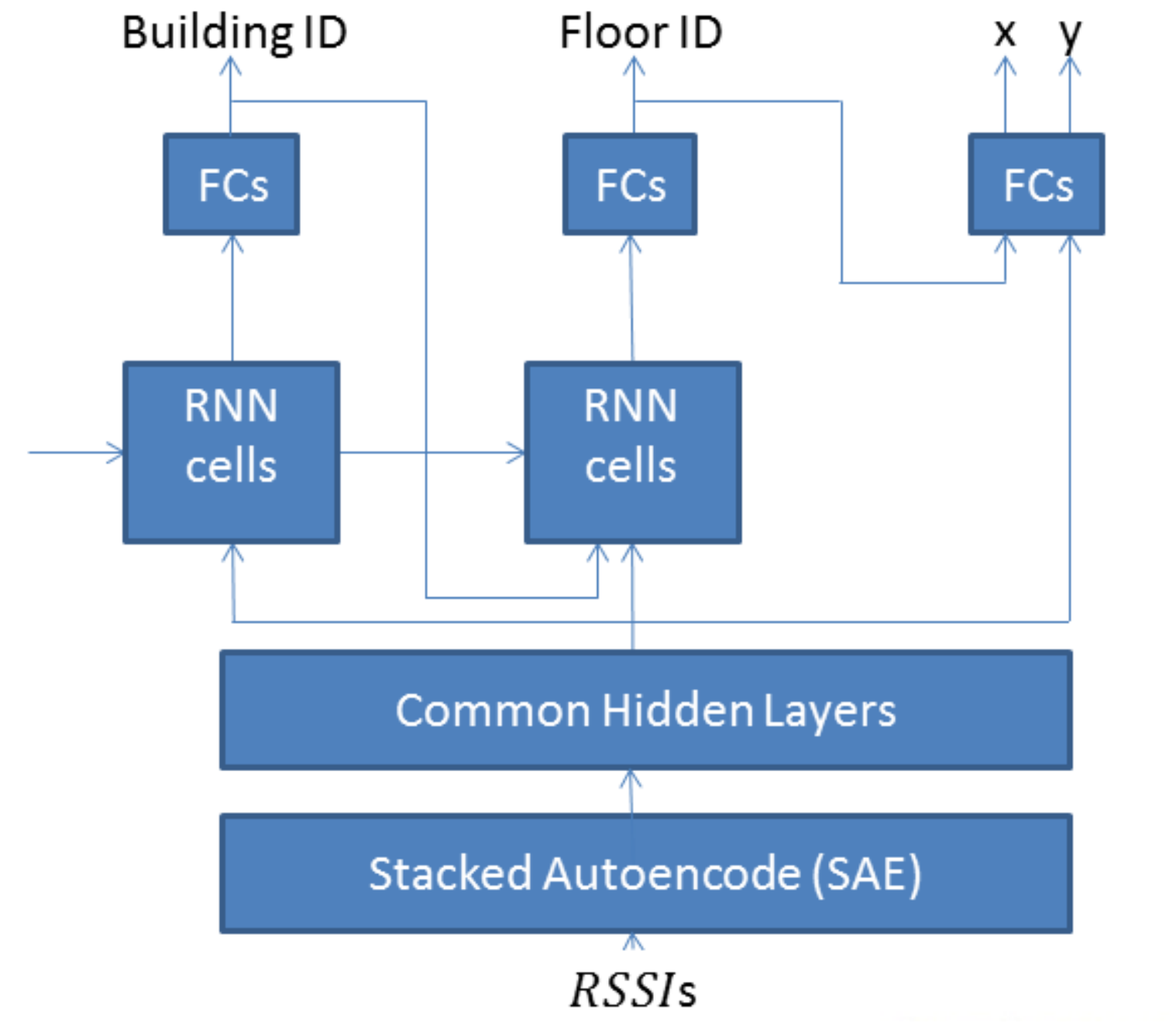}
    % {\scriptsize (b)}
  \end{center}
  \caption{Proposed network architecture based on RNN and SAE.}
  \label{fig:net-architecture}
\end{figure}
%%%
First, we fed the output from upper-level class to lower-level class to exploit
the hierarchical nature of the multi-building and multi-floor indoor
localization. Second, because the position estimation is different from building
and floor classification in nature, we exclude it from the RNN. Third, we add an
SAE before common hidden layers to reduce the dimensionality of a feature space
and thereby denoise RSSIs.
% The final model based on both RNN and SAE is shown in
% Fig.~\ref{fig:net-architecture}~(b).
% %%%
% \begin{figure}[!tb]
%   \centering%
%   \includegraphics[width=.65\linewidth]{modifiedrnn}
%   \caption{Network architecture of the final model based on RNN and SAE.}
%   \label{fig:modifiedrnn}
% \end{figure}
% % \Figure[!tb](topskip=0pt, botskip=0pt, midskip=0pt)[scale=0.7]{modifiedrnn.eps}
% % {Final modified structure.\label{fig:modifiedrnn}}
% %%%

% \subsection{Scalability}
% \label{sec:scalability}
% %%%
Note that scalability is one of the key challenges to be addressed in
large-scale multi-building and multi-floor indoor localization. Reducing the
number of output nodes is one of the major techniques to make the system
scalable. For the UJIIndoorLoc dataset, the number of output nodes would be 905
when we use multi-class classification, which could be reduced to 118---i.e.,
the sum of the number of buildings, the maximum number of floors, and the
maximum number of floor locations---by using multi-label
classification~\cite{Kim:18-2}. In the proposed work, we greatly reduce this
number to 4---i.e., 1 for building, 1 for floor, and 2 for location---by
numerical representation of outputs instead of one-hot-encoding. We convert
building and floor classification to regression problem and round the regression
outputs to get the class numbers.

\section{Experimental Results}
\label{sec:experimental-results}
%%%
We carry out experiments with the UJIIndoorLoc Wi-Fi fingerprinting dataset
\cite{torres-sospedra14:_ujiin} to investigate the effects of RNN parameter
values on the localization performance with a major focus on RNN cell types and
dropout rates. Note that the publicly-available UJIIndoorLoc dataset provides
only training and validation data; test data were provided only to the
competitors at the Evaluating Ambient Assisted Living (EvAAL)
competition~\cite{moreira15:_wi_fi}. Therefore, we split the training data into
new training and validation data with the ratio of 90:10, and we use the
validation data as test data.

As for performance metrics, we use classification accuracy and positioning
error: As for the classification accuracy, we calculate hit rates for building,
floor, and building/floor, the last of which counts only the correct
identification of both building and floor IDs; the positioning error is 2D or 3D
Euclidean distance between a predicted and true positions calculated as
discussed in \cite{moreira15:_wi_fi}.
% We also add 50 meters penalty for incorrect building estimation and 4 meters
% penalty for incorrect floor estimation to get 3D positioning error as in
% \cite{moreira15:_wi_fi}.

Table~\ref{tab:hyperparameters} summarizes the values of hyper parameters for
the experiments.
%%%
\begin{table}[!tb]
  \setlength\doublerulesep{0.15ex}%
  \caption{Hyper parameters}
  \label{tab:hyperparameters}
  \setlength{\tabcolsep}{3pt}
  \begin{center}
	\begin{tabular}{p{100pt} p{130pt}}
      \hline
      Parameter & Value \\
      \hline\hline
      SAE Hidden Layers & 256-128-64 \\
      SAE Activation  & ReLU \\
      SAE Optimizer & Adam \\
      SAE Loss & MSE \\
      \hline
      Common Hidden Layers & 128-128 \\
      Common Activation  & ReLU \\
      Common Dropout & 0.2 \\
      Common Loss & MSE \\
      \hline
      RNN Cells & 128-128 \\
      RNN Activation  & ReLU \\
      RNN Optimizer & Adam \\
      RNN Loss & MSE \\
      \hline
      BF Classifier Hidden Layers & 32-1 \\
      BF Classifier Activation  & MSE \\
      BF Classifier Optimizer & Adam \\
      BF Classifier Dropout & 0.2 \\
      BF Classifier Loss & ReLU \\
      \hline
      Position Hidden Layers & 128-128-2 \\
      Position Activation  & MSE \\
      Position Optimizer & Adam \\
      Position Dropout & 0.1 \\
      Position Loss & tanh \\
      \hline
	\end{tabular}
  \end{center}
\end{table}
%%%
We use SAE consists of three hidden layers of 256, 128, and 64 nodes, which is
mentioned as the best architecture for SAE in \cite{nowicki17:_low_wifi}. The
SAE is then followed by two common hidden layers with 128 nodes each. For
building and floor classifiers, we have two stacked RNN cells followed by two
fully-connected layers of 32 nodes and 1 node as the output node. Position
estimator consists of three fully-connected layers of 128, 128, and 2
nodes. Note that there are two output nodes for coordinates $(x, y)$. As for
epochs, we apply \textit{early stopping} with patience of 5, which forces early
stopping to run at least 5 epochs even though there is no improvement.

First, we compare the performance of two stacked RNN cell types, i.e., standard
RNN and long short-term memory (LSTM). The experiments were first conducted
using standard RNN cells over a range of the numbers of nodes, batch sizes, and
dropout rates and repeated again using LSTM cells, whose best results and
configurations are summarized in Tables~\ref{tab:resultsofcelltypes} and
\ref{tab:bestConfiguration}, respectively.
%%%
\begin{table}[!tb]
  \setlength\doublerulesep{0.15ex}%
  \caption{Results of different RNN cell types}
  \label{tab:resultsofcelltypes}
  \setlength{\tabcolsep}{3pt}
  \begin{center}
	\begin{tabular}{lccc}
      \hline
      RNN Cell Type & Building Hit Rate (\%) & Floor Hit Rate (\%) & Positioning
                                                                     Error (\si{\m}) \\
      \hline\hline
      Standard RNN & 100 & 94.42 & 8.68 \\ 
      \hline
      LSTM & 100 & 95.23 & 8.62 \\
      \hline
	\end{tabular}
  \end{center}
\end{table}
%%%
%%% 
\begin{table}[!tb]
  \setlength\doublerulesep{0.15ex}%
  \caption{Best configuration}
  \label{tab:bestConfiguration}
  \setlength{\tabcolsep}{3pt}
  \begin{center}
	\begin{tabular}{lc}
      \hline
      \multicolumn{1}{c}{Configuration} & Value \\
      \hline\hline
      RNN Cell Type & LSTM \\
      \hline
      Number of Nodes & 128 \\
      \hline
      Batch Size & 32 \\ 
      \hline
      Building/Floor Classifier Dropout & 0.2 \\
      \hline
      Building/Floor Classifier Epochs & 10 \\
      \hline
      Position Estimation Dropout & 0.1 \\
      \hline
      Position Estimation Epochs & 30 \\
      \hline
	\end{tabular}
  \end{center}
\end{table}
%%%
The results show that LSTM cell provides slightly better performance in floor
estimation and positioning error: The best results for standard RNN are 100\%
for building hit rate, 94.42\% for floor hit rate, and \SI{8.68}{\m} for
positioning error; for LSTM, the best results are 100\% for building hit rate,
95.23\% for floor hit rate, and \SI{8.62}{\m} for positioning
error. Table~\ref{tab:resultcomparison} compares our results against those of
other approaches based on the same UJIIndoorLoc data set, where we observe that
the proposed approach outperforms all other approaches except
CNNLoc~\cite{9060340}; CNNLoc, which requires a lot of data pre-processing
compared to the proposed one, shows slightly better floor hit rate of 96.03\%
but much higher positioning error of \SI{11.78}{\m}.
%%%
\begin{table}[!tb]
  \setlength\doublerulesep{0.15ex}%
  \caption{Comparison with other DNN-based approaches}
  \label{tab:resultcomparison}
  \setlength{\tabcolsep}{3pt}
  \begin{center}
	\begin{tabular}{lcc}
      \hline
      Approach & Floor Hit Rate (\%) & Positioning Error (\si{\m}) \\
      \hline\hline
      Proposed &  95.23 & 8.62 \\
      \hline
      DNN\cite{nowicki17:_low_wifi} & 92.00 & N/A \\
      \hline
      Scalable DNN\cite{Kim:18-1} & 91.27 & 9.29 \\
      \hline
      RF+SAE+Stacking\cite{9262763} & 95.13 & N/A \\
      \hline
      CNNLoc\cite{9060340} & 96.03 & 11.78 \\
      \hline
	\end{tabular}
  \end{center}
\end{table}
%%%

\begin{figure}[!tb]
  \centering%
  \includegraphics[width=.95\linewidth]{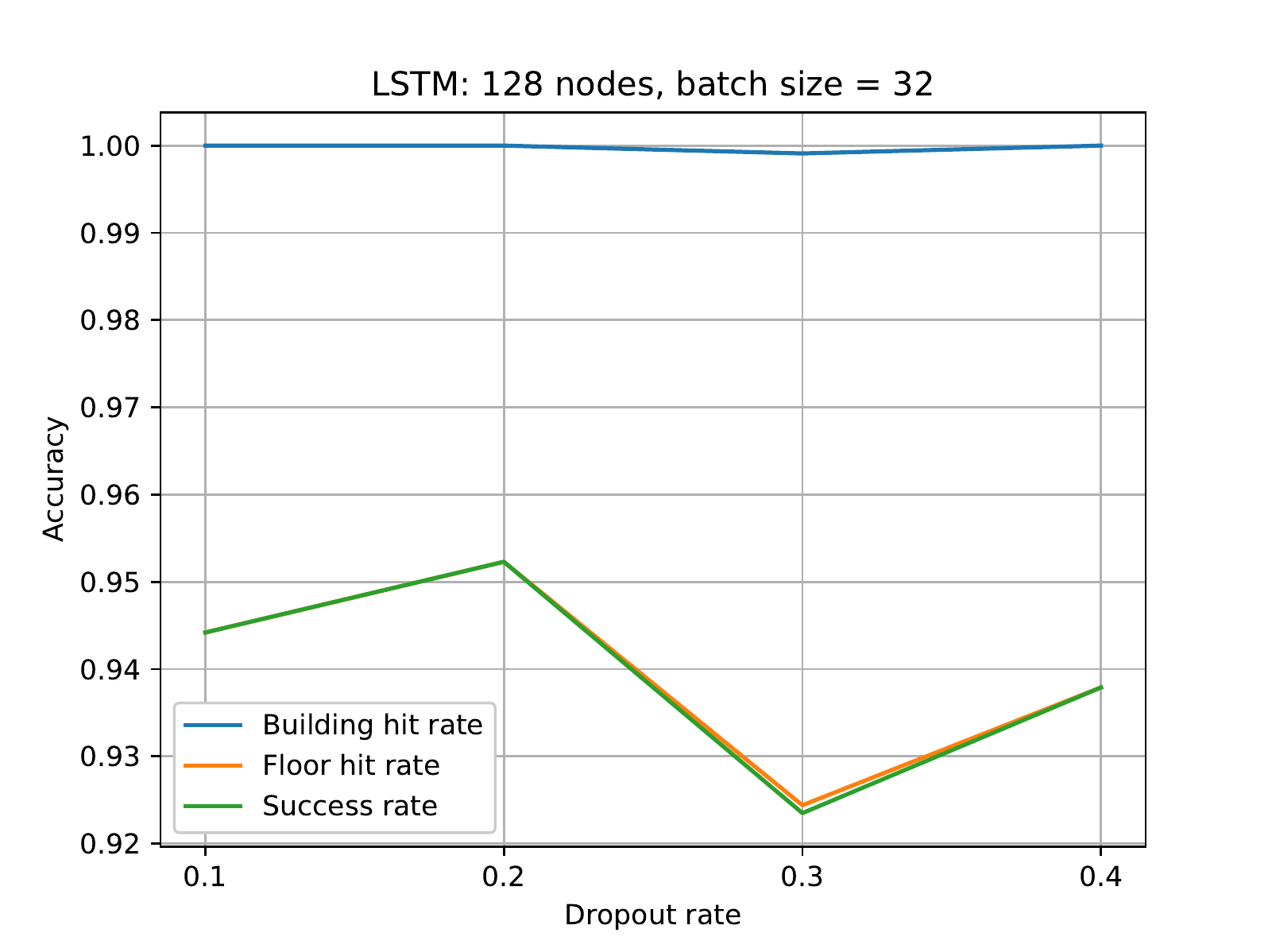}
  \caption{Effect of dropout rate on building and floor hit rate.}
  \label{fig:BFdropout}
\end{figure}

\begin{figure}[!tb]
  \centering%
  \includegraphics[width=.95\linewidth]{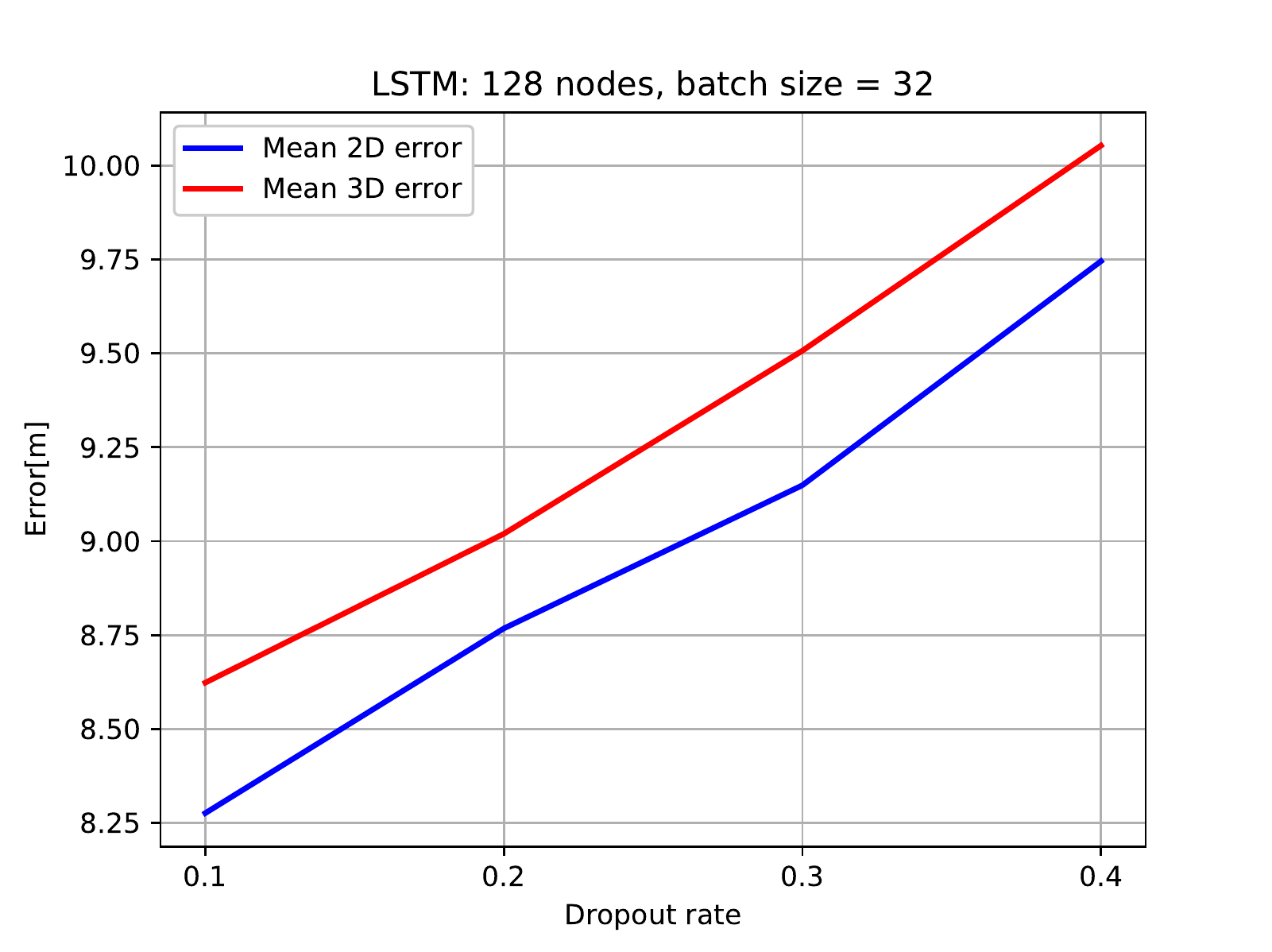}
  \caption{Effect of dropout rate on position estimation.}
  \label{fig:posdropout}
\end{figure}

We also investigate the effect of dropout rate on the localization performance
as shown in Figs.~\ref{fig:BFdropout} and \ref{fig:posdropout}.
%%%
% \Figure[!tb](topskip=0pt, botskip=0pt,
% midskip=0pt)[scale=0.5]{lstm_128_32_bfs.eps} {Different dropout rate for
%   building and floor hit rate.\label{fig:BFdropout}}
%%%
%%%
% \Figure[!tb](topskip=0pt, botskip=0pt,
% midskip=0pt)[scale=0.5]{lstm_128_32_err.eps} {Different dropout rate for
%   position estimation.\label{fig:posdropout}}
%%%
From the figures, we can observe that dropout rate of 0.2 gives the best results
for both building and floor estimation, while dropout rate of 0.1 always gives
the best results for coordinates estimation.

Table~\ref{tab:bestresults} compares our results against those of the best four
teams in the EvAAL competition.
%%%
\begin{table}[h]
  \setlength\doublerulesep{0.15ex}%
  \caption{Comparison with the results from EvAAL/IPIN 2015
    competition~\cite{moreira15:_wi_fi}}
  \label{tab:bestresults}
  \setlength{\tabcolsep}{3pt}
  \begin{center}
	\begin{tabular}{lccccc}
      \hline
      & Proposed & MOSAIC & HFTS & RTLSUM & ICSL \\
      \hline\hline
      Building Hit Rate (\%) & 100 & 98.65 & 100 & 100 & 100 \\
      \hline
      Floor Hit Rate (\%) & 95.23 & 93.86 & 96.25 & 93.74 &  86.93 \\
      \hline
      3D Positioning Error (\si{\m}) &8.62 & 11.64 & 8.49 & 6.20 & 7.67 \\
      \hline
	\end{tabular}
  \end{center}
\end{table}
%%%
Even though the objective and fair comparison is not possible due to the
unavailability of the original testing dataset, which were given only to the
participants of the EvAAL competition, the comparison in
Table~\ref{tab:bestresults} could give us an idea on the relative performance of
the proposed approach, where we find that the proposed approach outperforms
MOSAIC in all aspects and that our floor hit rate is higher than that of MOSAIC,
RTLSUM, and ICSL.

\section{Conclusions}
\label{sec:conclusions}
%%%
In this paper, we have proposed RNN-based hierarchical multi-building and
multi-floor indoor localization based on Wi-Fi fingerprinting. In our approach,
SAE and RNN are used for the reduction of feature space dimension and the
exploitation of the hierarchical nature of the localization in multi-building
and multi-floor environments, respectively.

Through the experimental results based on the publicly-available UJIIndoorLoc
dataset, we observe that the proposed indoor localization scheme achieves the
accuracy of 100\% and 95.23\% for building and floor estimation and 3D
positioning error of \SI{8.62}{\m}, which outperforms most of the existing
approaches including those based on DNNs.

Note that the proposed scheme clearly shows the advantages of hierarchical
indoor localization enabled by the use of RNN, while sharing the benefits of the
single-DNN-based schemes of \cite{Kim:18-1,Kim:18-2,Kim:18-3} like the
elimination of complicated data pre/post-processing and less parameter tuning.

\section*{Acknowledgment}
\label{sec:acknowledgment}
%%%
This work was supported in part by Postgraduate Research Scholarships (under
Grant PGRS1912001) and Key Program Special Fund (under Grant KSF-E-25) of Xi’an
Jiaotong-Liverpool University.

\balance % to balance the columns
%%

%%% References
% \bibliographystyle{IEEEtran}%%
% \bibliography{IEEEabrv,ae}%%

\end{document}